# Learning to Map Sentences to Logical Form: Structured Classification with Probabilistic Categorial Grammars

Luke S. Zettlemoyer and Michael Collins
MIT CSAIL
lsz@csail.mit.edu, mcollins@csail.mit.edu

## Abstract

This paper addresses the problem of mapping natural language sentences to lambda–calculus encodings of their meaning. We describe a learning algorithm that takes as input a training set of sentences labeled with expressions in the lambda calculus. The algorithm induces a grammar for the problem, along with a log-linear model that represents a distribution over syntactic and semantic analyses conditioned on the input sentence. We apply the method to the task of learning natural language interfaces to databases and show that the learned parsers outperform previous methods in two benchmark database domains.

## 1 Introduction

Recently, a number of learning algorithms have been proposed for *structured classification* problems. Structured classification tasks involve the prediction of output labels $y$ from inputs $x$ in cases where the output labels have rich internal structure. Previous work in this area has focused on problems such as sequence learning, where $y$ is a sequence of state labels (e.g., see (Lafferty, McCallum, & Pereira, 2001; Taskar, Guestrin, & Koller, 2003)), or natural language parsing, where $y$ is a context-free parse tree for a sentence $x$ (e.g., see Taskar et al. (2004)).

In this paper we investigate a new type of structured classification problem, where the goal is to learn to map natural language sentences to a lambda calculus encoding of their semantics. As one example, consider the following sentence paired with a logical form representing its meaning:

Sentence:    what states border texas
Logical Form: $\lambda x.state(x) \land borders(x, texas)$

The logical form in this case is an expression representing the set of entities that are states, and that also border Texas.

The training data in our approach consists of a set of sentences paired with logical forms, as in this example.

This is a particularly challenging problem because the derivation from each sentence to its logical form is not annotated in the training data. For example, there is no direct evidence that the word *states* in the sentence corresponds to the predicate *state* in the logical form; in general there is no direct evidence of the syntactic analysis of the sentence. Annotating entire derivations underlying the mapping from sentences to their semantics is highly labor-intensive. Rather than relying on full syntactic annotations, we have deliberately formulated the problem in a way that requires a relatively minimal level of annotation.

Our algorithm automatically induces a grammar that maps sentences to logical form, along with a probabilistic model that assigns a distribution over parses under the grammar. The grammar formalism we use is combinatory categorial grammar (CCG) (Steedman, 1996, 2000). CCG is a convenient formalism because it has an elegant treatment of a wide range of linguistic phenomena; in particular, CCG has an integrated treatment of semantics and syntax that makes use of a compositional semantics based on the lambda calculus. We use a log-linear model—similar to models used in conditional random fields (CRFs) (Lafferty et al., 2001)—for the probabilistic part of the model. Log-linear models have previously been applied to CCGs by Clark and Curran (2003), but our work represents a major departure from previous work on CCGs and CRFs, in that *structure learning* (inducing an underlying discrete structure, i.e., the grammar or CCG lexicon) forms a substantial part of our approach.

Mapping sentences to logical form is a central problem in designing natural language interfaces. We describe experimental results on two database domains: Geo880, a set of 880 queries to a database of United States geography; and Jobs640, a set of 640 queries to a database of job listings. Tang and Mooney (2001) described previous work on these data sets. Previous work by Thompson and Mooney (2002) and Zelle and Mooney (1996) used a subset of the Geo880 corpus. We evaluated the algorithm's accuracy in returning entirely correct logical forms for each test sentence. Our method achieves over 95% precision on both of these domains, with recall of 79% on each domain. These are highly competitive results when compared to the previous work.

## 2 Background

This section gives background material underlying our learning approach. We first describe the lambda–calculus expressions used to represent logical forms. We then describe combinatory categorial grammar (CCG), and the extension of CCG to probabilistic CCGs (PCCGs) through log-linear models.

### 2.1 Semantics

The sentences in our training data are annotated with expressions in a typed lambda–calculus language similar to the one presented by Carpenter (1997). The system has three basic types: $e$, the type of entities; $t$, the type of truth values; and $r$, the type of real numbers. It also allows functional types, for example $\langle e, t \rangle$, which is the type assigned to functions that map from entities to truth values. In specific domains, we will specify subtype hierarchies for $e$. For example, in a geography domain we might distinguish different entity subtypes such as cities, states, and rivers.

Figure 1 shows several sentences from the geography (Geo880) domain, together with their associated logical form. Each logical form is an expression from the lambda calculus. The lambda–calculus expressions we use are formed from the following items:

- **Constants:** Constants can either be entities, numbers or functions. For example, *texas* is an entity (i.e., it is of type $e$). *state* is a function that maps entities to truth values, and is of type $\langle e, t \rangle$. *size* is a function that maps entities to real numbers, and is therefore of type $\langle e, r \rangle$ (in the geography domain, *size(x)* returns the land-area of $x$).

- **Logical connectors:** The lambda–calculus expressions include conjunction ($\land$), disjunction ($\lor$), negation ($\neg$), and implication ($\rightarrow$).

- **Quantification:** The expressions include universal quantification ($\forall$) and existential quantification ($\exists$). For example, $\exists x.state(x) \land borders(x, texas)$ is true if and only if there is at least one state that borders Texas. Expressions involving $\forall$ take a similar form.

- **Lambda expressions:** Lambda expressions represent functions. For example, $\lambda x.borders(x, texas)$ is a function from entities to truth values, which is true of those states that border Texas.

**a)** What states border Texas
   $\lambda x.state(x) \land borders(x, texas)$

**b)** What is the largest state
   $\arg\max(\lambda x.state(x), \lambda x.size(x))$

**c)** What states border the state that borders the most states
   $\lambda x.state(x) \land borders(x, \arg\max(\lambda y.state(y),$
   $\quad \lambda y.count(\lambda z.state(z) \land borders(y, z))))$

Figure 1: Examples of sentences with their logical forms.

- **Additional quantifiers:** The expressions involve the additional quantifying terms *count*, $\arg\max$, $\arg\min$, and the definite operator $\iota$. An example of a count expression is $count(\lambda x.state(x))$, which returns the number of entities for which $state(x)$ is true. $\arg\max$ expressions are of the form $\arg\max(\lambda x.state(x), \lambda x.size(x))$. The first argument is a lambda expression denoting some set of entities; the second argument is a function of type $\langle e, r \rangle$. In this case the $\arg\max$ operator would return the set of items for which $state(x)$ is true, and for which $size(x)$ takes its maximum value. $\arg\min$ expressions are defined analogously. Finally, the definite operator creates expressions such as $\iota(\lambda x.state(x))$. In this case the argument is a lambda expression denoting some set of entities. $\iota(\lambda x.state(x))$ would return the unique item for which $state(x)$ is true, if a unique item exists. If no unique item exists, it causes a presupposition error.

### 2.2 Combinatory Categorial Grammars

The parsing formalism underlying our approach is that of combinatory categorial grammar (CCG) (Steedman, 1996, 2000). A CCG specifies one or more logical forms—of the type described in the previous section—for each sentence that can be parsed by the grammar.

The core of any CCG is a lexicon, $\Lambda$. In a purely syntactic version of CCG, the entries in $\Lambda$ consist of a word (lexical item) paired with a syntactic type. A simple example of a CCG lexicon is as follows:

$$
\begin{array}{lll}
\text{Utah} & := & NP \\
\text{Idaho} & := & NP \\
\text{borders} & := & (S \backslash NP)/NP
\end{array}
$$

In this lexicon *Utah* and *Idaho* have the syntactic type $NP$, and *borders* has the more complex type $(S \backslash NP)/NP$. A syntactic type can be either one of a number of *primitive categories* (in the example, $NP$ or $S$), or it can be a *complex type* of the form $A/B$ or $A \backslash B$ where both $A$ and $B$ can themselves be a primitive or complex type. The primitive categories $NP$ and $S$ stand for the linguistic notions of noun-phrase and sentence respectively. Note that a single word can have more than one syntactic type, and hence more than one entry in the lexicon.

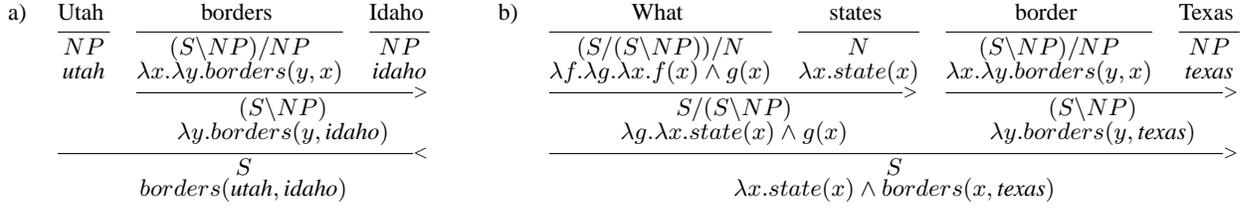

Figure 2: Two examples of CCG parses.

In addition to the lexicon, a CCG has a set of *combinatory rules* which describe how adjacent syntactic categories in a string can be recursively combined. The simplest such rules are rules of *functional application*, defined as follows:

(1) *The functional application rules:*
   a. $A/B \quad B \quad \Rightarrow \quad A$
   b. $B \quad A\backslash B \quad \Rightarrow \quad A$

Intuitively, a category of the form $A/B$ denotes a string that is of type $A$ but is missing a string of type $B$ to its right; similarly, $A\backslash B$ denotes a string of type $A$ that is missing a string of type $B$ to its left.

The first rule says that a string with type $A/B$ can be combined with a right-adjacent string of type $B$ to form a new string of type $A$. As one example, in our lexicon, *borders*, (which has the type $(S\backslash NP)/NP$) can be combined with *Idaho* (which has the type $NP$) to form the string *borders Idaho* with type $S\backslash NP$. The second rule is a symmetric rule applying to categories of the form $A\backslash B$. We can use this to combine *Utah* (type $NP$) with *borders Idaho* (type $S\backslash NP$) to form the string *Utah borders Idaho* with the type $S$. We can draw a parse tree (or derivation) of *Utah borders Idaho* as follows:

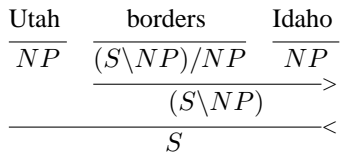

Note that we use the notation $\rightarrow$ and $\prec$ to denote application of rules 1(a) and 1(b) respectively.

CCGs typically include a *semantic type*, as well as a syntactic type, for each lexical entry. For example, our lexicon would be extended as follows:

Utah $\quad := \quad NP : utah$
Idaho $\quad := \quad NP : idaho$
borders $\quad := \quad (S\backslash NP)/NP : \lambda x.\lambda y.borders(y, x)$

We use the notation $A : f$ to describe a category with syntactic type $A$ and semantic type $f$. Thus *Utah* now has syntactic type $NP$, and semantic type $utah$. The functional application rules are then extended as follows:

(2) *The functional application rules (with semantics):*
   a. $A/B : f \quad B : g \quad \Rightarrow \quad A : f(g)$
   b. $B : g \quad A\backslash B : f \quad \Rightarrow \quad A : f(g)$

Rule 2(a) now specifies how the semantics of the category $A$ is compositionally built out of the semantics for $A/B$ and $B$. Our derivations are then extended to include a compositional semantics. See Figure 2(a) for an example parse. This parse shows that *Utah borders Idaho* has the syntactic type $S$ and the semantics *borders(utah, idaho)*.

In spite of their relative simplicity, CCGs can capture a wide range of syntactic and semantic phenomena. As one example, see Figure 2(b) for a more complex parse. Note that in this case we have an additional primitive category, $N$ (for nouns), and the final semantics is a lambda expression denoting the set of entities that are states and that border Texas. In this case, the lexical item *what* has a relatively complex category, which leads to the correct analysis of the underlying string.

A full description of CCG goes beyond the scope of this paper. There are several extensions to the formalism: see (Steedman, 1996, 2000) for more details. In particular, CCG includes rules of combination that go beyond the simple function application rules in 1(a) and 1(b).[1] Additional combinatory rules allow CCGs to give an elegant treatment of linguistic phenomena such as coordination and relative clauses. In our work we make use of standard rules of application, forward and backward composition, and type-raising. In addition, we allow lexical entries consisting of strings of length greater than one, for example

the Mississippi $\quad := \quad NP : mississippi\_river$

This leads to a relatively minor change to the formalism, which in practice can be very useful. For example, it is easier to directly represent the fact that *the Mississippi* refers to the Mississippi river with the lexical entry above than it is to try to construct this meaning compositionally from the meanings of the determiner *the* and the word *Mississippi*, which refers to the state of Mississippi when used without the determiner.

---
[1]One example of a more complex combinatory rule is that of *forward composition*:
$$A/B : f \quad B/C : g \Rightarrow A/C : \lambda x.f(g(x))$$
Another rule which is frequently used is that of *type-raising*:
$$NP : f \quad \Rightarrow \quad NP/(S\backslash NP) : \lambda g.g(f)$$
This would allow *Utah* to be type-raised to a category $NP/(S\backslash NP) : \lambda g.g(Utah)$.

### 2.3 Probabilistic CCGs

We now describe how to generalize CCGs to probabilistic CCGs (PCCGs). A CCG, as described in the previous section, will generate one or more derivations for each sentence $S$ that can be parsed by the grammar. We will describe a derivation as a pair $(L, T)$, where $L$ is the final logical form for the sentence (e.g., $borders(utah, idaho)$ in figure 2(a)), and $T$ is the sequence of steps taken in deriving $L$. We will frequently refer to $T$ as a parse tree. A PCCG defines a conditional distribution $P(L, T|S)$ over possible $(L, T)$ pairs for a given sentence $S$.

In general, various sources of ambiguity can lead to a sentence $S$ having more than one valid $(L, T)$ pair. This is the primary motivation for extending CCGs to PCCGs: PCCGs deal with ambiguity by ranking alternative parses for a sentence in order of probability. One source of ambiguity is lexical items having more than one entry in the lexicon. For example, *New York* might have entries $NP : new\_york\_city$ and $NP : new\_york\_state$. Another source of ambiguity is where a single logical form $L$ may be derived by multiple derivations $T$. This latter form of ambiguity can occur in CCG, and is often referred to as *spurious ambiguity*; the term *spurious* is used because the different syntactic parses lead to identical semantics.

In defining PCCGs, we make use of a conditional log-linear model that is similar to the model form in conditional random fields (CRFs) (Lafferty et al., 2001) or log-linear models applied to parsing (Ratnaparkhi, Roukos, & Ward, 1994; Johnson, Geman, Canon, Chi, & Riezler, 1999). Log-linear models for CCGs are described in (Clark & Curran, 2003). We assume a function $\bar{f}$ mapping $(L, T, S)$ triples to feature vectors in $\mathbb{R}^d$. This function is defined by $d$ individual features, so that $\bar{f}(L, T, S) = \langle f_1(L, T, S), \ldots, f_d(L, T, S) \rangle$. Each feature $f_j$ is typically the count of some sub-structure within $(L, T, S)$. The model is parameterized by a vector $\bar{\theta} \in \mathbb{R}^d$. The probability of a particular (syntax, semantics) pair is defined as

$$P(L, T|S; \bar{\theta}) = \frac{e^{\bar{f}(L,T,S) \cdot \bar{\theta}}}{\sum_{(L,T)} e^{\bar{f}(L,T,S) \cdot \bar{\theta}}} \quad (1)$$

The sum in the denominator is over all valid parses for $S$ under the CCG grammar.

In this paper we make use of *lexical features* alone. For each lexical entry in the grammar, we have a feature $f_j$ that counts the number of times that the lexical entry is used in $T$. For example, in the simple grammar with entries for *Utah*, *Idaho* and *borders*, there would be three features of this type. While these features are quite simple, we have found them to be quite successful when applied to the Geo880 and Jobs640 data sets. More complex features are certainly possible (e.g., see (Clark & Curran, 2003)). In the future, we would like to explore more general features that have been shown to be useful in other parsing settings.

### 2.4 Parsing and Parameter Estimation

We now turn to issues of parsing and parameter estimation. Parsing under a PCCG involves computing the most probable logical form $L$ for a sentence $S$,

$$\arg\max_L P(L|S; \bar{\theta}) = \arg\max_L \sum_T P(L, T|S; \bar{\theta})$$

where the $\arg\max$ is taken over all logical forms $L$ and the hidden syntax $T$ is marginalized out by summing over all parses that produce $L$. We use dynamic programming algorithms for this step, which are very similar to CKY–style algorithms for parsing probabilistic context-free grammars (PCFGs).[2] Dynamic programming is feasible within our approach because the feature-vector definitions $\bar{f}(L, T, S)$ involve *local* features that keep track of counts of lexical items in the derivation $T$.[3]

In parameter estimation, we assume that we have $n$ training examples, $\{(S_i, L_i) : i = 1 \ldots n\}$. $S_i$ is the $i$'th sentence in training data, and $L_i$ is the lambda expression associated with that sentence. The task is to estimate the parameter values $\bar{\theta}$ from these examples. Note that the training set does not include derivations $T_i$, and we therefore view derivations as hidden variables within the approach. The log-likelihood of the training set is given by:

$$\begin{aligned} O(\bar{\theta}) &= \sum_{i=1}^n \log P(L_i|S_i; \bar{\theta}) \\ &= \sum_{i=1}^n \log \left( \sum_T P(L_i, T|S_i; \bar{\theta}) \right) \end{aligned}$$

Differentiating with respect to $\theta_j$ yields:

$$\begin{aligned} \frac{\partial O}{\partial \theta_j} &= \sum_{i=1}^n \sum_T f_j(L_i, T, S_i) P(T|S_i, L_i; \bar{\theta}) \\ &\quad - \sum_{i=1}^n \sum_{L,T} f_j(L, T, S_i) P(L, T|S_i; \bar{\theta}) \end{aligned}$$

The two terms in the derivative involve the calculation of expected values of a feature under the distributions

---

[2]CKY–style algorithms for PCFGs (Manning & Schutze, 1999) are related to the Viterbi algorithm for hidden Markov models, or dynamic programming methods for Markov random fields.

[3]We use beam–search during parsing, where low-probability sub-parses are discarded at some points during parsing, in order to improve efficiency.

$P(T|S_i, L_i; \bar{\theta})$ or $P(T, L|S_i; \bar{\theta})$. Expectations of this type can again be calculated using dynamic programming, using a variant of the inside-outside algorithm (Baker, 1979), which was originally formulated for probabilistic context-free grammars.

Given this derivative, we can use it directly to maximize the likelihood using a stochastic gradient ascent algorithm (LeCun, Bottou, Bengio, & Haffner, 1998),[4] which takes the following form:

Set $\bar{\theta}$ to some initial value
**for** $k = 0 \ldots N - 1$
   **for** $i = 1 \ldots n$
      $\bar{\theta} = \bar{\theta} + \frac{\alpha_0}{(1+ct)} \frac{\partial \log P(L_i|S_i; \bar{\theta})}{\partial \bar{\theta}}$

where $t = i + k \times n$ is the total number of previous updates, $N$ is a parameter that controls the number of passes over the training data, and $\alpha_0$ and $c$ are learning–rate parameters.

## 3 Learning

In the previous section we saw that a probabilistic Combinatory Categorial Grammar (PCCG) is defined by a lexicon $\Lambda$, together with a parameter vector $\bar{\theta}$. In this section, we present an algorithm that learns a PCCG. One input to the algorithm is a training set of $n$ examples, $\{(S_i, L_i) : i = 1 \ldots n\}$, where each training example is a string $S_i$ paired with a logical form $L_i$. Another input to the algorithm is an initial lexicon, $\Lambda_0$.[5]

Note that the training data includes neither direct evidence about the parse trees mapping each $S_i$ to $L_i$, nor the set of lexical entries which are required for this mapping. We treat the parse trees as a hidden variable within our model. The set of possible parse trees for a sentence depends on the lexicon, which is itself learned from the training examples. Thus, at a high level, learning will involve the following two sub-problems:

- Induction of a lexicon, $\Lambda$, which defines a set of parse trees for each training sentence $S_i$.

- Estimation of parameter values, which define a distribution over parse trees for any sentence.

[4]The EM algorithm could also be used, but would require some form of gradient ascent for the M–step. Because of this, we found it simpler to use gradient ascent for the entire optimization.

[5]In our experiments the initial lexicon includes lexical items that are derived directly from the database in the domain; for example, we have a list of entries $\{Utah := NP : utah, Idaho := NP : idaho, Nevada := NP : nevada, \ldots\}$ including every U.S. state in the geography domain. It also includes lexical items that are domain independent, and easily specified by hand: for example, the definition for "what" in Figure 2(b) would be included, as it would be useful across many domains.

The first problem can be thought of as a form of *structure learning*, and is a major focus of the current section. The second problem is a more conventional *parameter estimation* problem, which roughly speaking can be solved using the gradient descent methods described in section 2.4.

The remainder of this section describes an overall strategy for these two problems. We show how to interleave a structure-building step, GENLEX, with a parameter estimation step, in a way that results in a PCCG with a compact lexicon and effective parameter estimates for the weights of the log-linear model. Section 3.1 describes the main structural step, GENLEX$(S, L)$, which generates a set of candidate lexical items that may be useful in deriving $L$ from $S$. In section 3.2 we describe the overall learning algorithm, which prunes the lexical entries suggested by GENLEX and estimates the parameters of a log-linear model.

### 3.1 Lexical Learning

We now describe the function GENLEX, which takes a sentence $S$ and a logical form $L$ and generates a set of lexical items. Our aim is to define GENLEX$(S, L)$ in such a way that the set of lexical items that it generates allows at least one parse of $S$ that results in $L$.

As an example, consider the parse in Figure 2(a). When presented with the input sentence *Utah borders Idaho* and logical form $borders(utah, idaho)$, we would like GENLEX to produce a lexicon that includes the three lexical items that were used in this parse, namely

| Utah | := | $NP : utah$ |
| Idaho | := | $NP : idaho$ |
| borders | := | $(S \backslash NP)/NP : \lambda x.\lambda y.borders(y, x)$ |

Our definition of GENLEX will also produce *spurious* lexical items, such as $borders := NP : idaho$ and $borders\ utah := (S\backslash NP)/NP : \lambda x.\lambda y.borders(y, x)$. Later, we will see how these items can be pruned from the lexicon in a later stage of processing.

To compute GENLEX, we make use of a function, $C(L)$, that maps a logical form to a set of categories (such as $NP : utah$, or $NP : idaho$). GENLEX is then defined as

$$\text{GENLEX}(S, L) = \{x := y \mid x \in W(S), y \in C(L)\}$$

where $W(S)$ is the set of all subsequences of words in $S$.

The function $C(L)$ is defined through a set of rules that examine $L$ and produce categories based on its structure. Figure 3 shows the rules that we use. Each rule consists of a *trigger* that identifies some sub-structure within the logical form $L$. For each sub-structure in $L$ that matches the trigger, a category is created and added to $C(L)$. As one example, the second row in the table defines a rule that identifies all arity-one predicates $p$ within the logical form as triggers for creating a category $N : \lambda x.p(x)$. Given the logical form $\lambda x.major(x) \wedge city(x)$, which has the arity-one predicates $major$ and $city$, this rule would create the categories $N : \lambda x.major(x)$ and $N : \lambda x.city(x)$.

| | Rules | Categories produced from logical form |
|---|---|---|
| Input Trigger | Output Category | $\arg\max(\lambda x.state(x) \wedge borders(x, texas), \lambda x.size(x))$ |
| constant $c$ | $NP : c$ | $NP : texas$ |
| arity one predicate $p_1$ | $N : \lambda x.p_1(x)$ | $N : \lambda x.state(x)$ |
| arity one predicate $p_1$ | $S \backslash NP : \lambda x.p_1(x)$ | $S \backslash NP : \lambda x.state(x)$ |
| arity two predicate $p_2$ | $(S \backslash NP)/NP : \lambda x.\lambda y.p_2(y, x)$ | $(S \backslash NP)/NP : \lambda x.\lambda y.borders(y, x)$ |
| arity two predicate $p_2$ | $(S \backslash NP)/NP : \lambda x.\lambda y.p_2(x, y)$ | $(S \backslash NP)/NP : \lambda x.\lambda y.borders(x, y)$ |
| arity one predicate $p_1$ | $N/N : \lambda g.\lambda x.p_1(x) \wedge g(x)$ | $N/N : \lambda g.\lambda x.state(x) \wedge g(x)$ |
| literal with arity two predicate $p_2$ and constant second argument $c$ | $N/N : \lambda g.\lambda x.p_2(x, c) \wedge g(x)$ | $N/N : \lambda g.\lambda x.borders(x, texas) \wedge g(x)$ |
| arity two predicate $p_2$ | $(N \backslash N)/NP : \lambda x.\lambda g.\lambda y.p_2(x, y) \wedge g(x)$ | $(N \backslash N)/NP : \lambda g.\lambda x.\lambda y.borders(x, y) \wedge g(x)$ |
| an arg max / min with second argument arity one function $f$ | $NP/N : \lambda g.\arg\max/\min(g, \lambda x.f(x))$ | $NP/N : \lambda g.\arg\max(g, \lambda x.size(x))$ |
| an arity one numeric-ranged function $f$ | $S/NP : \lambda x.f(x)$ | $S/NP : \lambda x.size(x)$ |

Figure 3: The rules that define GENLEX. We use the term *predicate* to refer to a function that returns a truth value; *function* to refer to all other functions; and *constant* to refer to constants of type $e$. Each row represents a rule. The first column lists the triggers that identify some sub-structure within a logical form $L$, and then generate a category. The second column lists the category that is created. The third column lists example categories that are created when the rule is applied to the logical form at the top of this column.

Intuitively, each of the rules in Figure 3 corresponds to a different linguistic sub-category such as noun, transitive verb, adjective, and so on. For example, the rule in the first row generates categories that are noun phrases, and the second rule generates nouns. The end result is an efficient way to generate a large set of linguistically plausible categories $C(L)$ that could be used to construct a logical form $L$.

### 3.2 The Learning Algorithm

Figure 4 shows the learning algorithm used within our approach. The output of the algorithm is a PCCG, defined by a lexicon $\Lambda$ and a parameter vector $\bar{\theta}$. As input, the algorithm takes a training set of sentences paired with logical forms, together with an initial lexicon, $\Lambda_0$.

At all stages, the algorithm maintains a parameter vector $\bar{\theta}$ which stores a real value associated with every possible lexical item. The set of possible lexical items is

$$\Lambda^* = \Lambda_0 \cup \bigcup_{i=1}^{n} \text{GENLEX}(S_i, L_i)$$

In our experiments, the parameters were initialized to be 0.1 for all lexical items in $\Lambda_0$, and 0.01 for all other lexical items. These values were chosen through experiments on the development data; they give a small initial bias towards using lexical items from $\Lambda_0$ and favor parses that include more lexical items.

The goal of the algorithm is to provide a relatively compact lexicon, which is a small subset of the entire set of possible lexical items. The algorithm achieves this by alternating between two steps. The goal of step 1 is to search for a relatively small number of lexical entries, which are nevertheless sufficient to successfully parse all training examples. Step 2 is then used to re-estimate the parameters of the lexical items that are selected in step 1.

In the $t$'th application of step 1, each sentence in turn is parsed with the current parameters $\bar{\theta}^{t-1}$ and a special, sentence–specific lexicon which is defined as $\Lambda_0 \cup \text{GENLEX}(S_i, L_i)$. This will result in one or more highest-scoring parses that have the logical form $L_i$.[6] Lexical items are extracted from these highest-scoring parses alone. The result of this stage is to generate a small subset $\lambda_i$ of $\text{GENLEX}(S_i, L_i)$ for each training example. The output of step 1, at iteration $t$, is a subset of $\Lambda^*$, defined as $\Lambda_t = \Lambda_0 \cup \bigcup_{i=1}^{n} \lambda_i$.

Step 2 re-estimates the parameters of the members of $\Lambda_t$, using stochastic gradient descent. The starting point for gradient descent when estimating $\bar{\theta}^t$ is $\bar{\theta}^{t-1}$, i.e., the parameter values at the previous iteration. For any lexical item that is not a member of $\Lambda_t$, the associated parameter in $\bar{\theta}^t$ is set to be the same as the corresponding parameter in $\bar{\theta}^{t-1}$ (i.e., parameter values are simply copied across from the previous iteration).

The motivation for cycling between steps 1 and 2 is as follows. In step 1, keeping only those lexical items that occur in the highest scoring parse(s) leading to $L_i$ results in a compact lexicon. This step is also guaranteed to produce a lexicon $\Lambda_t \subset \Lambda^*$ such that the accuracy on the training data when running the PCCG $(\Lambda_t, \bar{\theta}^{t-1})$ is at least as accurate as applying the PCCG $(\Lambda^*, \bar{\theta}^{t-1})$. In other words, pruning the lexicon in this way cannot hurt parsing performance on training data in comparison to using all possible lexical entries.[7]

---

[6] Note that this set of highest-scoring parses is identical to the set produced by parsing with $\Lambda^*$, rather than the sentence-specific lexicon. This is because $\Lambda_0 \cup \text{GENLEX}(S_i, L_i)$ contains all lexical items that can possibly be used to derive $L_i$.

[7] To see this, note that restricting the lexicon in this way cannot exclude any of the highest-scoring parses for $S_i$ that lead to $L_i$. In practice, it may exclude some parses that lead to logical forms for $S_i$ that are incorrect. Because the highest-scoring correct parses

Step 2 also has a guarantee, in that the log-likelihood on the training data will improve (assuming that stochastic gradient descent is successful in improving its objective). Step 2 is needed because after each application of step 1, the parameters $\bar{\theta}^{t-1}$ are optimized for $\Lambda_{t-1}$ rather than $\Lambda_t$, the current lexicon. Step 2 derives new parameter values $\bar{\theta}_t$ which are optimized for $\Lambda_t$.

In summary, steps 1 and 2 together form a greedy, iterative method for simultaneously finding a compact lexicon and also optimizing the log-likelihood of the model on the training data.

## 4 Related Work

This section discusses related work on natural language interfaces to databases (NLIDBs), in particular focusing on learning approaches, and related work on learning CCGs.

There has been a significant amount of work on hand engineering NLIDBs. Androutsopoulos, Ritchie, and Thanisch (1995) provide a comprehensive summary of this work. Recent work in this area has focused on improved parsing techniques and designing grammars that can be ported easily to new domains (Popescu, Armanasu, Etzioni, Ko, & Yates, 2004).

Zelle and Mooney (1996) developed one of the earliest examples of a learning system for NLIDBs. This work made use of a deterministic shift–reduce parser and developed a learning algorithm, called CHILL, based on techniques from Inductive Logic Programming, to learn control rules for parsing. The major limitation of this approach is that it does not learn the lexicon, instead assuming that a lexicon that pairs words with their semantic content (but not syntax) has been created in advance. Later, Thompson and Mooney (2002) developed a system that learns a lexicon for CHILL that performed almost as well as the original system. Most recently, Tang and Mooney (2001) developed a statistical shift–reduce parser that significantly outperformed these original systems. However, this system, again, does not learn a lexicon.

A number of previous learning methods (Papineni, Roukos, & Ward, 1997; Ramaswamy & Kleindienst, 2000; Miller, Stallard, Bobrow, & Schwartz, 1996; He & Young, 2004) have been applied to the ATIS domain, which involves a natural language interface to a travel database of flight information. In the future we plan to test our method on this domain. Miller et al. (1996) describe an approach that assumes full annotation of parse trees. Papineni et al. (1997) and Ramaswamy and Kleindienst (2000) use approaches based on methods originally developed for machine translation. He and Young (2004) describe an approach using an extension of hidden Markov models, resulting in a model with some of the power of context-free models.

are still allowed, parsing performance cannot deteriorate.

**Inputs:**
- Training examples $E = \{(S_i, L_i) : i = 1 \ldots n\}$ where each $S_i$ is a sentence, each $L_i$ is a logical form.
- An initial lexicon $\Lambda_0$

**Procedures:**
- PARSE$(S, L, \Lambda, \bar{\theta})$: takes as input a sentence $S$, a logical form $L$, a lexicon $\Lambda$, and a parameter vector $\bar{\theta}$. Returns the highest probability parse for $S$ with logical form $L$, when $S$ is parsed by a PCCG with lexicon $\Lambda$ and parameters $\bar{\theta}$. If there is more than one parse with the same highest probability, the entire set of highest probability parses is returned. Dynamic programming methods are used when implementing PARSE, see section 2.4 of this paper.
- ESTIMATE$(\Lambda, E, \bar{\theta})$: takes as input a lexicon $\Lambda$, a training set $E$, and a parameter vector $\bar{\theta}$. Returns parameter values $\bar{\theta}$ that are the output of stochastic gradient descent on the training set $E$ under the grammar defined by $\Lambda$. The input $\bar{\theta}$ is the initial setting for the parameters in the stochastic gradient descent algorithm. Dynamic programming methods are used when implementing ESTIMATE, see section 2.4.
- GENLEX$(S, L)$: takes as input a sentence $S$ and a logical form $L$. Returns a set of lexical items. See section 3.1 for a description of GENLEX.

**Initialization:** Define $\bar{\theta}$ to be a real-valued vector of arity $|\Lambda^*|$, where $\Lambda^* = \Lambda_0 \cup \bigcup_{i=1}^n \text{GENLEX}(S_i, L_i)$. $\bar{\theta}$ stores a parameter value for each potential lexical item. The initial parameters $\bar{\theta}^0$ are taken to be 0.1 for any member of $\Lambda_0$, and 0.01 for all other lexical items.

**Algorithm:**
- For $t = 1 \ldots T$

    **Step 1:** (Lexical generation)
    - For $i = 1 \ldots n$:
        – Set $\lambda = \Lambda_0 \cup \text{GENLEX}(S_i, L_i)$.
        – Calculate $\pi = \text{PARSE}(S_i, L_i, \lambda, \bar{\theta}^{t-1})$.
        – Define $\lambda_i$ to be the set of lexical entries in $\pi$.
    - Set $\Lambda_t = \Lambda_0 \cup \bigcup_{i=1}^n \lambda_i$

    **Step 2:** (Parameter Estimation)
    - Set $\bar{\theta}^t = \text{ESTIMATE}(\Lambda_t, E, \bar{\theta}^{t-1})$

**Output:** Lexicon $\Lambda_T$ together with parameters $\bar{\theta}^T$.

Figure 4: The overall learning algorithm.

There have been several pieces of previous work on learning CCGs. Clark and Curran (2003) developed a method for leaning the parameters of a log-linear model for syntactic CCG parsing given fully annotated normal–form parse trees. Watkinson and Manandhar (1999) presented an unsupervised approach for learning CCGs that, again, does not perform any semantic analysis. We know of only one previous system (Bos, Clark, Steedman, Curran, & Hockenmaier, 2004) that learns CCGs with semantics. However, this approach requires fully–annotated CCG derivations as supervised training data. As such, the techniques they employed are not applicable to learning in our framework.

|  | Geo880 | | Jobs640 | |
| --- | --- | --- | --- | --- |
|  | P | R | P | R |
| Our Method | 96.25 | 79.29 | 97.36 | 79.29 |
| COCKTAIL | 89.92 | 79.40 | 93.25 | 79.84 |

Figure 5: The results for our method, and the previous work of COCKTAIL, when applied to the two database query domains. $P$ is precision in recovering entire logical forms, $R$ is recall.

## 5 Experiments

We evaluated the learning algorithm on two domains: Geo880, a set of 880 queries to a database of U.S. geography; and Jobs640, a set of 640 queries to a database of job listings. The data were originally annotated with Prolog style semantics which we manually converted to equivalent statements in the lambda calculus.

We compare the structured classifier results to the COCKTAIL system (Tang & Mooney, 2001). The COCKTAIL experiments were conducted by performing ten–fold cross validation of the entire data set. We used a slightly different experimental set-up, where we made an explicit split between training and test data sets.[8] The Geo880 data set was divided into 600 training examples and 280 test examples; the Jobs640 set was divided into 500 training and 140 test examples. The parameters of the training algorithm were tuned by cross–validation on the training set. We did two passes of the overall learning loop in Figure 4. Each time we used gradient descent to estimate parameters, we performed three passes over the training set with the learning-rate parameters $\alpha_0 = 0.1$ and $c = 0.001$.

We give precision and recall for the different algorithms, defined as $Precision = \# \ correct/total \ \# \ parsed$, $Recall = \# \ correct/total \ \# \ examples$. Sentences are correct if the parser gives a completely correct semantics.

Figure 5 shows the results of the experiments. Our approach has higher precision than COCKTAIL on both domains, with a very small reduction in recall. When evaluating these results, it is important to realize that COCKTAIL is provided with a fairly extensive lexicon that pairs words with semantic predicates. For example, the word *borders* would be paired with the predicate $borders(x, y)$. This prior information goes substantially beyond the initial lexicon used in our own experiments.[9]

---

[8]This allowed us to use cross-validation experiments on the *training set* to optimize parameters, and more importantly to develop our algorithms while ensuring that we had not implicitly tuned our approach to the final test set.

[9]Note that the work of (Thompson & Mooney, 2002) does describe a method which automatically learns a lexicon. However, results for this approach were worse than results for CHILL (Zelle & Mooney, 1996), which in turn were considerably worse than results for COCKTAIL on the Geo250 domain, a subset of the examples in Geo880.

| states | := | $N : \lambda x.state(x)$ |
| --- | --- | --- |
| major | := | $N/N : \lambda f.\lambda x.major(x) \wedge f(x)$ |
| population | := | $N : \lambda x.population(x)$ |
| cities | := | $N : \lambda x.city(x)$ |
| rivers | := | $N : \lambda x.river(x)$ |
| run through | := | $(S \backslash NP)/NP : \lambda x.\lambda y.traverse(y, x)$ |
| the largest | := | $NP/N : \lambda f. \arg\max(f, \lambda x.size(x))$ |
| river | := | $N : \lambda x.river(x)$ |
| the highest | := | $NP/N : \lambda f. \arg\max(f, \lambda x.elev(x))$ |
| the longest | := | $NP/N : \lambda f. \arg\max(f, \lambda x.len(x))$ |

Figure 6: Ten learned lexical items that had highest associated parameter values from a randomly chosen development run in the Geo880 domain.

To better understand these results, we examined performance of our method through cross-validation on the training set. We found that the approach creates a compact lexicon for the training examples that it parses. On the Geo880 domain, the initial number of lexical items created by GENLEX was on average 393.8 per training example After pruning, on average only 5.1 lexical items per training example remained. The Jobs640 domain showed a reduction from an average of 697.1 lexical items per training example, to 6.6 items.

To investigate the disparity between precision and recall, we examined the behavior of the algorithm when trained in the cross-validation (development) regime. We found that on average, the learner failed to parse 9.3% of the training examples in the Geo880 domain, and 8.7% of training examples in the Jobs640 domain. (Note that sentences which cannot be parsed in step 1 of the training algorithm are excluded from the training set during step 2.) These parse failures were caused by sentences whose semantics could not be built from the lexical items that GENLEX created. For example, the learner failed to parse complex sentences such as *Through which states does the Mississippi run* because GENLEX does not create lexical entries that allow the verb *run* to find its argument, the preposition *through*, when it has moved to the front of the sentence. This problem is almost certainly a major cause of the lower recall on test examples. Exploring the addition of more rules to GENLEX is an important area for future work.

Figure 6 gives a sample of lexical entries that are learned by the approach. These entries are linguistically plausible and should generalize well to unseen data.

## 6 Discussion and Future Work

In this paper, we presented a learning algorithm that creates accurate structured classifiers for natural language interfaces. A major focus for future work is to apply the algorithm to a range of larger data sets. Larger data sets should improve the recall performance and allow us to develop a more comprehensive set of rules for GENLEX, ultimately creating a robust system that can quickly learn interfaces for new, unseen domains with little human assistance.

Although the experiments in this paper only learned natural language interfaces to databases, there are many other natural language interfaces that the techniques can be generalized to handle. In particular, we will explore building interfaces to dialogue systems. These interfaces must handle a much wider range of semantic phenomena (for example, anaphora and ellipses). Extending the current algorithm to address these challenges will greatly increase the range of possible interfaces that are successfully learned.

**Acknowledgements**

We would like to thank Rohit Kate and Raymond Mooney for their help with obtaining the Geo880 and Jobs640 data sets. We also gratefully acknowledge the support of a ND-SEG graduate research fellowship and the National Science Foundation under grants 0347631 and 0434222.